\documentclass{article}
\usepackage{ijcai25}

\usepackage{times}
\usepackage{soul}
\usepackage{url}
\usepackage[hidelinks]{hyperref}
\usepackage[utf8]{inputenc}
\usepackage[small]{caption}
\usepackage{graphicx}
\usepackage{amsmath}
\usepackage{amsthm}
\usepackage{amssymb}
\usepackage{booktabs}
\usepackage{enumitem}
\usepackage{algorithm}
\usepackage{algorithmic}
\usepackage[switch]{lineno}
\usepackage{multirow}
\usepackage{colortbl}  
\usepackage{xcolor}    
\usepackage{tikz}

\def\revision{\textcolor{black}}

\newtheorem{definition}{Definition}

\title{Generative Co-Design of Antibody Sequences and Structures via Black-Box Guidance in a Shared Latent Space}

\author{
Yinghua Yao$^{1,2}$\and
Yuangang Pan$^{1,2}$\thanks{Corresponding author}\and
Xixian Chen$^{3}$
\\
\affiliations
$^1$Center for Frontier AI Research, Agency for Science, Technology and Research, Singapore\\
$^2$Institute of High Performance Computing, Agency for Science, Technology and Research, Singapore\\
$^3$Singapore Institute of Food and Biotechnology Innovation, Agency for Science, Technology and Research, Singapore
\emails
{eva.yh.yao, yuangang.pan}@gmail.com,
xixian\_chen@sifbi.a-star.edu.sg
}

\begin{document}

\maketitle

\begin{abstract}
Advancements in deep generative models have enabled the joint modeling of antibody sequence and structure, given the antigen-antibody complex as context. However, existing approaches for optimizing complementarity-determining regions (CDRs) to improve developability properties operate in the raw data space, leading to excessively costly evaluations due to the inefficient search process. To address this, we propose LatEnt blAck-box Design (LEAD), a sequence-structure co-design framework that optimizes both sequence and structure within their shared latent space. Optimizing shared latent codes can not only break through the limitations of existing methods, but also ensure synchronization of different modality designs. Particularly, we design a black-box guidance strategy to accommodate real-world scenarios where many property evaluators are non-differentiable. Experimental results demonstrate that our LEAD achieves superior optimization performance for both single and multi-property objectives. Notably, LEAD reduces query consumption by a half while surpassing baseline methods in property optimization. The code is available at \url{https://github.com/EvaFlower/LatEnt-blAck-box-Design}.
\end{abstract}

\section{Introduction}

Antibodies are glycoproteins produced by the immune system that specifically bind to antigens—foreign molecules that trigger immune responses. Antigen recognition and binding arise primarily from the complementarity-determining regions (CDRs) of antibodies~\cite{janeway2001immunobiology,presta1992antibody}. As such, the design of CDRs is a pivotal step in developing effective therapeutic antibodies and plays a key role in advancing drug discovery.

The space of possible antibodies is vast and discrete, and experimental validation is slow and costly~\cite{larman2012construction}. This drives a significant development of \textit{in silico} antibody design approaches to help build a small, enriched library of candidates to identify viable options with minimal experimental measurements.
Traditional computational methods sample antibody sequences/structures based on hand-crafted and statistical energy functions~\cite{lapidoth2015abdesign,adolf2018rosettaantibodydesign,warszawski2019optimizing}, which are inefficient and easily trapped in local optima. Recent advancements in diffusion-based deep generative models enable the simultaneous distribution modeling of both the sequence and structure of CDRs~\cite{jin2022iterative,luo2022antigen,kong2023end}, thereby improving design efficiency. These models have the advantage of incorporating both antigen epitopes and antibody framework structures during the design process, which is essential to enhance binding affinity.

The clinical effectiveness of antibody-based therapeutics relies not only on their biological performance—such as target specificity and potency—but also on their biophysical and chemical characteristics, collectively referred to as ``developability''. Critical aspects of developability, including solubility, thermal stability, aggregation tendency, and immunogenicity, must be carefully optimized to guarantee both manufacturability and safety in clinical applications~\cite{Amelia2024guiding,sumida2024improving}. While most efforts primarily focus on enhancing antigen-binding properties, some recent works~\cite{Amelia2024guiding,kulyte2024improving,zhou2024antigen} incorporate developability into antibody design by guiding either the sequence or structure in the raw data space, or fine-tuning the model, which usually requires extensive property evaluations. 

\begin{figure*}[!t]
\centering 
\centerline{\includegraphics[width=.9\textwidth]{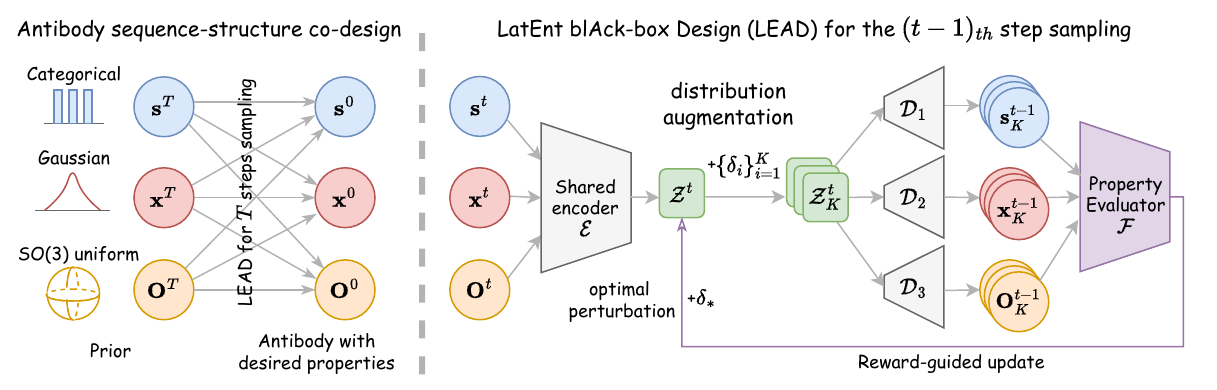}}\vskip-.1in
\caption{\label{fig:framework} Overall framework. \textbf{Left panel} describes the whole process of our guided sampling. \textbf{Right panel} details the black-box guidance incorporated in the shared latent space of sequence and structure.}\vskip-.1in
\end{figure*}

In this work, we employ deep generative models that learn a joint antibody sequence and structure distribution to design antibody CDRs de novo, optimizing not only for antigen affinity but also for enhanced developability. Following the principle that sequence determines structure, and structure determines function~\cite{branden2012introduction}, we propose a novel LatEnt blAck-box Design (LEAD) framework (Fig.~\ref{fig:framework}), which guides property optimization through the shared latent space of sequences and structures. Specifically, to address real-world scenarios where many property evaluators are not differentiable, we design our guidance strategy based on black-box techniques, allowing flexible and general-purpose optimization across diverse antibody design tasks.
The contributions are summarized as follows:
\begin{itemize}
\item We propose LatEnt blAck-box Design (LEAD), a method that jointly optimizes sequence and structure within the shared latent space of a diffusion model for antigen-specific antibody design. LEAD establishes a general framework, and different variants of LEAD are proposed for single and multiple property optimization.
\item LEAD enables property guidance by incorporating black-box optimization into the denoising process, without any additional training of diffusion models.
\item Experimental results demonstrate that LEAD achieves superior optimization efficiency compared to existing approaches that perform guidance in the raw antibody space. Specifically, it requires fewer property evaluations while maintaining or exceeding performance in both single and multiple-property optimization.
\end{itemize}

\section{Related work}
\paragraph{Antibody Design} Deep learning methods have achieved great success in antibody design. Large Language Model (LLM)-based approaches are efficient in generating protein sequences~\cite{saka2021antibody,shuai2023iglm}, but they struggle to design antibodies for specific antigens since protein function is usually shaped by a complex interplay between sequence and structural features. 
The first CDR sequence-structure co-design model is introduced for designing antibodies against SARS-CoV-2~\cite{jin2022iterative}, but its reliance on an antigen-specific predictor limits its generalizability to other antigens. 
Recently, diffusion models have shown significant potential in both continuous and categorical data domains and have been successfully applied to generating high-quality protein sequences or protein backbones~\cite{watson2023novo,ingraham2023illuminating,lisanza2024multistate,gruver2024protein}.
DiffAb~\cite{luo2022antigen} proposed the first diffusion model to perform joint design of sequence and structure of the antibody CDR regions while conditioning on the antigen-antibody complex. AbDiffuser~\cite{martinkus2024abdiffuser} improved this further by incorporating strong priors and being more memory efficient with side chain generation. DiffAbXL~\cite{ucar2024exploring} is an advanced version of DiffAb, with an improved architecture.

\paragraph{Property Optimization} Sequence-structure co-design methods enhance the antigen-targeting performance of antibodies. Recent studies further incorporate property optimization to improve their developability. DIFFFORCE~\cite{kulyte2024improving} introduces a differentiable force field w.r.t. the 3D atom coordinates to guide the sampling process. ABDPO~\cite{zhou2024antigen} fine-tunes DiffAb via direct preference optimization to optimize energies of generated antibodies. 
\cite{Amelia2024guiding} performs guidance in the raw space of antibodies according to developability properties during diffusion sampling. However, these methods require extensive property evaluations, limiting their scalability and efficiency.

\section{Black-box antibody design in the shared latent space of diffusion model}
In this section, we first introduce the background of diffusion model for antibody sequence and structure co-design. Then we develop our black-box guidance strategy in the shared latent space of this co-design model.

\subsection{Diffusion model for antibody co-design}\label{joint_diffusion}

Our work builds upon the diffusion model for antibody sequence and structure co-design. It enables the joint generation of Complementarity-Determining Region (CDR) sequences and structures conditioning on the antibody framework and its corresponding bound antigen. Specifically, each amino acid is represented by its type $s_i\in \textrm{\{ACDEFGHIKLMNPQRSTVWY\}}$, $\mathrm{C}_\alpha$ atom coordinate $x_i \in \mathbb{R}^3$, and amino acid orientation $O_i \in \mathrm{SO}(3)$, where $i=1,2, \ldots, N$ and $N$ is the total number of the amino acids in the  antibody-antigen complex. 

The target CDR loop is denoted as $\mathcal{A} = \{\left(s_i, x_i, O_i\right) |i=1, 2,\ldots, m\}$, given the rest of the antibody-antigen complex $\mathcal{R}=\{\left(s_j, x_j, O_j\right) |j\ne i, j\in \{1, 2,\ldots, N\}\}$. For the sake of notation, let $\mathbf{s}=\{s_i\}_{i=1}^m, \mathbf{x}=\{x_i\}_{i=1}^m, \mathbf{O}=\{O_i\}_{i=1}^m$ denote the whole feature of each modality of the target CDR loop, respectively. Then, we have $\mathcal{A} =  \left(\mathbf{s}, \mathbf{x}, \mathbf{O}\right)$. The antibody generation task can be then formulated as modeling the conditional distribution $P(\mathcal{A} |\mathcal{R})$. DiffAb~\cite{luo2022antigen} represents the first application of Denoising Diffusion Probabilistic Model (DDPM)~\cite{ho2020denoising} to antibody generation, which consists of a forward diffusion process and a reverse generative process. 

\paragraph{Forward diffusion process $(t=1,\ldots, T)$}
The antibody sequence-structure co-design framework initially assumes the three modalities are conditionally independent. Namely,
\begin{equation} \label{diffusion_independent}
    q(\mathcal{A}^t|\mathcal{A}^0) = q\left(\mathbf{s}^t |\mathbf{s}^0\right)q\left(\mathbf{x}^t |\mathbf{x}^0\right)q\left(\mathbf{O}^t |\mathbf{O}^0\right),
\end{equation}
where $\mathcal{A}^0 = \left(\mathbf{s}^0=\{s_i^0\}_{i=1}^m, \mathbf{x}^0=\{x_i^0\}_{i=1}^m, \mathbf{O}^0=\{O_i^0\}_{i=1}^m\right)$ is the noisy-free CDR loop at time step~0. Similarly, $\mathcal{A}^t = \left(\mathbf{s}^t, \mathbf{x}^t, \mathbf{O}^t\right)$ is the noisy CDR loop at time step $t$.

Then, the forward diffusion process gradually introduces noise into each modality through different distributions towards the prior distributions. Namely,
\begin{subequations}
\begin{align}
& q\left(s^t_i |s^0_i\right)=\mathcal{C}\left(\mathbb{I}(s^t_i) |\bar{\alpha}^t \mathbb{I}(s^0_i)+\bar{\beta}^t \mathbf{1}/20\right),\\
& q\left(x^t_i |x^0_i\right)=\mathcal{N}\left(x^t_i |\sqrt{\bar{\alpha}^t} x^0_i, \bar{\beta}^t \mathbf{I}\right),\\
& q\left(O^t_i |O^0_i\right)=\mathcal{I} \mathcal{G}_{\text {SO(3) }}\left(O^t_i |\textrm{ScaleRot}
(\sqrt{\bar{\alpha}_t}, O^0_i), \bar{\beta}^t\right),
\end{align}
\end{subequations}
where $i=1,2,\ldots, m$ is the index for the amino acid. $\mathbb{I}(\cdot)$ is the one-hot operation that converts each amino acid type to a 20-dimensional one-hot vector and $\mathbf{1}$ is an all-one vector. $\{\beta^t\}_{t=1}^T$ is the noise schedule for the diffusion process, and we define $\bar{\alpha}^t=\prod_{\tau=1}^t(1-\beta^\tau)$ and $\bar{\beta}^t=1-\bar{\alpha}^t$. Here, $\mathcal{C}(\cdot), \mathcal{N}(\cdot)$, and $\mathcal{I} \mathcal{G}_{\mathrm{SO}(3)}(\cdot)$ are the categorical distribution, Gaussian distribution on $\mathbb{R}^3$, and isotropic Gaussian distribution on SO(3)~\cite{leach2022denoising} respectively. ScaleRot scales the rotation angle with a fixed rotation axis to modify the rotation matrix~\cite{gallier2003computing}.

\paragraph{Reverse generative process $(t=T,T-1,\ldots, 1)$}\label{Reverse_generative_process}
The reverse generative diffusion process learns to recover data by iteratively denoising from the prior distributions. The parametric models $p_\theta\left(\mathcal{A}^{t-1} |\mathcal{A}^t, \mathcal{R}\right)$ are employed to approximate the posterior distributions at each generation time step. To model the interplay between sequences and structures, a three-output neural network for the three modalities with a shared encoder and separate decoders~\cite{luo2022antigen} is adopted, which takes the whole sequence-structure context $\mathcal{R}$ and the noisy antibody at the previous denoising step $\mathcal{A}^t=(\mathbf{s}^t, \mathbf{x}^t, \mathbf{O}^t)$ as input and predicts $\mathcal{A}^{t-1}=\left(\mathbf{s}^{t-1}, \mathbf{x}^{t-1}, \mathbf{O}^{t-1}\right)$. Namely,
\begin{align}\label{denoising_independent}
    p_\theta\left(\mathcal{A}^{t-1} |\mathcal{A}^t, \mathcal{R}\right) = p_{\theta}\left(\mathbf{s}^{t-1} |\mathcal{A}^t, \mathcal{R}\right)
    &p_{\theta}\left(\mathbf{x}^{t-1} |\mathcal{A}^t, \mathcal{R}\right) \\
    \times &p_{\theta}\left(\mathbf{O}^{t-1} |\mathcal{A}^t, \mathcal{R}\right).\nonumber
\end{align}
In particular, the denoising process of three modalities from time step $t$ to time step $t-1$ can be defined as follows:
\begin{subequations}\label{Generation_process}
\begin{align}
& p_{\theta}\left(s^{t-1}_i |\mathcal{A}^t, \mathcal{R}\right)=\mathcal{C}\left(\mathbb{I}(s^{t-1}_i) |f_{1,i}\left(\mathcal{A}^t, \mathcal{R}\right)\right),\\
& p_{\theta}\left(x^{t-1}_i |\mathcal{A}^t, \mathcal{R}\right)=\mathcal{N}\left(x^{t-1}_i |f_{2,i}\left(\mathcal{A}^t, \mathcal{R}\right), \beta^t \mathbf{I}\right), \\
& p_{\theta}\left(O^{t-1}_i |\mathcal{A}^t, \mathcal{R}\right)=\mathcal{I} \mathcal{G}_{\mathrm{SO}(3)}\left(O^{t-1}_i|f_{3,i}\left(\mathcal{A}^t, \mathcal{R}\right), \beta^t\right),
\end{align}
\end{subequations}
where $i=1,2,\ldots, m$ is the index for the amino acid. $f_{1}, f_{2}, f_{3}$ are parameterized by $\mathrm{SE}(3)$-equivariant neural networks, whose $i$-th entry denotes the output for the $i$-th amino acid, respectively.

\subsection{Function-guided latent optimization}

With the pre-trained joint diffusion model defined in Section~\ref{joint_diffusion}, we explore how to incorporate the functional guidance $\mathcal{F}(\mathcal{A}^t)$ (without loss of generality, we assume higher function reward is better), defined with respect to sequence and structure, into the denoising sampling process to generate antibodies with targeted properties.

To enable more efficient functional guidance, we no longer implement guidance in the raw data space. Instead, we consider the shared latent space of sequences and structures. 

First of all, we define the shared latent space of multiple modalities.  According to the definition of the three-output denoising neural network, we reformulate it as follows:
\begin{subequations}\label{shared_encoder}
\begin{align}
& f_{1}\left(\mathcal{A}^t, \mathcal{R}\right) = \mathcal{D}_{1}\left(\mathcal{E}\left(\mathcal{A}^t, \mathcal{R}\right)\right)= \mathcal{D}_{1}(\mathcal{Z}^t), \\
& f_{2}\left(\mathcal{A}^t, \mathcal{R}\right) = \mathcal{D}_{2}\left(\mathcal{E}\left(\mathcal{A}^t, \mathcal{R}\right)\right)= \mathcal{D}_{2}(\mathcal{Z}^t), \\
& f_{3}\left(\mathcal{A}^t, \mathcal{R}\right) = \mathcal{D}_{3}\left(\mathcal{E}\left(\mathcal{A}^t, \mathcal{R}\right)\right)= \mathcal{D}_{3}(\mathcal{Z}^t),
\end{align}
\end{subequations}
where $\mathcal{E}(\cdot)$ denotes the shared encoder while $\mathcal{D}_{*}(\cdot)$ denotes the modality-specific decoder. Denoting $\mathcal{E}\left(\mathcal{A}^t, \mathcal{R}\right)$ as $\mathcal{Z}^t$, we consider the space where $\mathcal{Z}^t$ resides as the shared latent space of multiple modalities. 

To avoid extra noise distraction,  we adopt the deterministic version of the reverse generative process, a.k.a. Denoising Diffusion Implicit Models (DDIM)~\cite{song2020denoising}. For simplicity of notation, let DDIM also denote a one-step reverse generative process consisting of a three-output denoising network with a shared encoder~($\mathcal{E}, \mathcal{D}_1, \mathcal{D}_2, \mathcal{D}_3$). 

\begin{definition}\label{DDIM}[DDIM for deterministic sampling]
Given the noisy antibody $\mathcal{A}^t$ and the rest antigen-antibody context $\mathcal{R}$, the subsequent denoised antibody $\mathcal{A}^{t-1}$ can be obtained deterministically via DDIM, i.e., $\mathcal{A}^{t-1} \sim DDIM\left(\mathcal{D}_1, \mathcal{D}_2, \mathcal{D}_{3}, \mathcal{Z}^t, \beta_t\right)$ where $\mathcal{Z}^t=\mathcal{E}\left(\mathcal{A}^t, \mathcal{R}\right)$ and $\beta_t$ is the $t$-step noise schedule as in Eq.~\eqref{Generation_process}. For simplicity, we denote $\mathcal{A}^{t-1}=\mathcal{G}(\mathcal{Z}^t), t=1, 2, \ldots, T$.
\end{definition}

In order to obtain high reward design, we propose optimizing the shared latent code~$\mathcal{Z}^t$ to obtain the high-reward denoising CDR loop~$\mathcal{A}^{t-1}=\mathcal{G}(\mathcal{Z}^t)$ at each sampling step. Namely,
\begin{equation}\label{eq:opt_z}
    \max_{\mathcal{Z}^t} \mathcal{F}\left(\mathcal{G}(\mathcal{Z}^t)\right), \forall \; t=T, T-1, \ldots, 1.
\end{equation}
\revision{Since the shared embedding $\mathcal{Z}^t$ encodes both sequence and structure, any updates to$\mathcal{Z}^t$ affect all three modalities simultaneously, ensuring their synchronization throughout the optimization process.}


\begin{algorithm}[tb]
        \caption{LatEnt blAck-box Design (LEAD)}\label{Guided_DiffAb}
        \begin{algorithmic}[1]
        \STATE {\bfseries Input:} a pre-trained antibody diffusion model featuring three-output denosing network with shared encoder~($\mathcal{E}, \mathcal{D}_{1}, \mathcal{D}_{2}, \mathcal{D}_{3}$), the antibody-antigen backbone~$\mathcal{R}$, the dynamic step size~$\{\beta_t\}_{t=1}^T$, the property evaluator~$\mathcal{F} (\uparrow)$, number of evaluations per time step $K$, initial guidance step~$T_{\text{init}}<=T$, the step size $\delta$.
        \STATE {\bfseries Initialize:} $\mathcal{A}^T= \big(\mathbf{s}^T\sim \mathcal{C}\left(\mathbb{I}(\mathbf{s}^T) |\mathbf{1}/M \right), \mathbf{x}^T\sim \mathcal{N}(\mathbf{x}^T|\mathbf{0}, \mathbf{I}), \mathbf{O}^T \sim \mathcal{I} \mathcal{G}_{\mathrm{SO}(3)}\left(\mathbf{I}, 1\right)\big)$.
        \FOR {$t = T,T-1,\ldots, 1$}
          \IF{$t>T_{\text{init}}$}
          \STATE $\mathcal{A}^{t-1} \sim DDPM\left(\mathcal{D}_{1}, \mathcal{D}_{2}, \mathcal{D}_{3}, \mathcal{E}\left(\mathcal{A}^t, \mathcal{R}\right),\beta_t\right)$;
          \ELSE
          \STATE $\mathcal{Z}^t = \mathcal{E}\left(\mathcal{A}^t, \mathcal{R}\right)$;
          \FOR {$k = 1,2,\ldots, K$}
          \STATE $\mathcal{Z}^t_k = \mathcal{Z}^t + \sigma\delta^t_k$ where $\delta^t_k \sim \mathcal{N}\left(\mathbf{0}, \mathbf{I}\right)$;
          \STATE $\mathcal{A}^{t-1}_k \sim DDIM\left(\mathcal{D}_{1}, \mathcal{D}_{2}, \mathcal{D}_{3}, \mathcal{Z}^t_k, \beta_t\right)$;
         \ENDFOR
         \STATE Obtain the optimal noise $\delta^t_{\star}$ based on Section~\ref{reward_guided};
          \STATE $\mathcal{A}^{t-1}_{\star} \sim DDIM\left(\mathcal{D}_{1}, \mathcal{D}_{2}, \mathcal{D}_{3}, \mathcal{Z}^t + \sigma\delta^t_{\star}, \beta_t \right)$;
          \ENDIF
        \ENDFOR
        \STATE {\bfseries Output:} generated CDR loop $\mathcal{A}^0_{\star}$.
        \end{algorithmic}
\end{algorithm}   

\subsection{Reward-guided updates}
\label{reward_guided}
Most property evaluators and denoising sampling for the CDRs are not differentiable, so we cannot perform white-box functional guidance. Motivated by the black-box optimization technique~\cite{wierstra2014natural,nesterov2017random}, we propose learning the distribution of latent codes~$\mathcal{Z}^t$ to enable black-box gradient updates, thereby facilitating an efficient optimization of~$\mathcal{Z}^t$. Let $\pi(\mathcal{Z}^t; \mathcal{W}^t)$ denote the distribution of latent code, where $\mathcal{W}^t$ is the distribution parameter. We can reformulate Eq.~\eqref{eq:opt_z} as
\begin{equation}\label{eq:opt_z_bo}
\begin{aligned}
    \max_{\pi(\mathcal{Z}^t;\mathcal{W}^t)} J(\mathcal{W}^t)&=\mathbb{E}_{\mathcal{Z}^t\sim \pi(\mathcal{Z}^t;\mathcal{W}^t)}\left[\mathcal{F}\left(\mathcal{G}(\mathcal{Z}^t)\right)\right],\\
    \forall \; t&=T, T-1, \ldots, 1.
\end{aligned}
\end{equation}
Using the log-likelihood trick, the Monte Carlo gradient estimation w.r.t. $\mathcal{W}^t$ is estimated as:
\begin{equation}\label{eq:bo_grad}
\nabla_{\mathcal{W}^t} J(\mathcal{W}^t) \approx \frac{1}{K} \sum_{k=1}^K \mathcal{F} \left( \mathcal{G}(\mathcal{Z}_k^t) \right) \nabla_{\mathcal{W}^t} \log \pi(\mathcal{Z}_k^t; \mathcal{W}^t),
\end{equation}
where $\mathcal{Z}_k^t$ is sampled from $\pi(\mathcal{Z}^t;\mathcal{W}^t)$. $K$ is the number of function queries used to refine $\mathcal{W}^t$.

To further simplify, we instantiate the population distribution $\pi$ as an isotropic multivariate Gaussian with mean $\mathcal{W}^t$ and fixed covariance $\sigma^2\mathbf{I}$, allowing us to write $J(\mathcal{W}^t)$ in terms of $\mathcal{W}^t$ directly. Denoting $\mathcal{W}^t=\mathcal{Z}^t=\mathcal{E}(\mathcal{A}^t, \mathcal{R})$, Eq.~\eqref{eq:opt_z_bo} can be reformulated as:
\begin{equation}\label{new_obj}
\begin{aligned}
    \max_{\mathcal{Z}^t} J(\mathcal{Z}^t)&=\mathbb{E}_{\delta^t\sim \mathcal{N}(\mathbf{0}, \mathbf{I})}\left[\mathcal{F}\left(\mathcal{G}(\mathcal{Z}^t+\sigma\delta^t)\right)\right],\\
    \forall \; t&=T, T-1, \ldots, 1.
\end{aligned}    
\end{equation}
Based on Eq.~\eqref{eq:bo_grad} and Eq.~\eqref{new_obj}, we obtain the gradient w.r.t. $\mathcal{Z}^t$:
\begin{equation}\label{eq:bo_grad_}
\nabla_{\mathcal{Z}^t} \mathbb{E} \left[ \mathcal{F} \left( \mathcal{G}(\mathcal{Z}^t) \right) \right] \approx \frac{1}{\sigma K} \sum_{k=1}^K \delta_k^t \mathcal{F} \left( \mathcal{G}(\mathcal{Z}^t + \sigma \delta_k^t) \right).
\end{equation}
Thus, the latent code $\mathcal{Z}^t$ can be updated as follows:
\begin{equation}\label{eq:bo_update}
    \mathcal{Z}^t \leftarrow \mathcal{Z}^t+\frac{1}{\sigma K} \sum_{k=1}^K \delta_k^t \mathcal{F} \left( \mathcal{G}(\mathcal{Z}^t + \sigma \delta_k^t)\right).
\end{equation}
However, we find that Eq.~\eqref{eq:bo_update} fails to achieve an effective optimization in the case of sequence derived properties (see empirical study). We hypothesize that it may result from the discrete nature of sequence space. Inspired from the fact that Eq.~\eqref{eq:bo_update} actually updates $\mathcal{Z}^t$ with a reward-weighted noise, we propose sampling an optimal noise based on probability derived from function rewards to be compatible with the discrete nature of data. Specifically, the $K$ reward values can be converted into probabilities via the softmax function:
\begin{equation}
    P_k^t = \frac{\exp\left(\mathcal{F}\left(\mathcal{G}(\mathcal{Z}^t+\sigma\delta_k^t)\right)\right)}
{\sum_{j=1}^{K} \exp\left(\mathcal{F}\left(\mathcal{G}(\mathcal{Z}^t+\sigma\delta_j^t)\right)\right)}.
\end{equation}
The optimized noise index is sampled from this distribution:
\begin{equation}\label{soft_sel}
    \gamma^t \sim \text{Categorical}(P_1^t, P_2^t, \dots, P_K^t),
\end{equation}
which is a one-hot vector. The optimized noise can be derived by $\delta^t_{\star} = \sum_{j=1}^K \gamma^t_j \delta^t_j$. The latent code is updated by:
\begin{equation}
    \mathcal{Z}^t \leftarrow \mathcal{Z}^t+\sigma\delta^t_{\star}.
\end{equation}
It is intuitive to derive a hard selection version of the above strategy by defining:
\begin{equation}\label{eq:best_n_latent}
    \delta^t_{\star} = \arg\max_{\delta^t_k} \{\mathcal{F}\left(\mathcal{G}(\mathcal{Z}^t+\sigma\delta_k^t)\right)\}_{k=1}^{K}.
\end{equation}
We refer to this strategy as the hard selection version of LEAD, termed LEAD-H, and to Eq.~\eqref{soft_sel} as the soft selection version of LEAD, termed LEAD-S.

Additionally, we find that combining LEAD-H/LEAD-S with the weighted version of LEAD (termed LEAD-W) yields better results, respectively. We hypothesize that LEAD-W can explore a larger search space, thereby enhancing the advantages of LEAD-H/LEAD-S. The combining strategy is as the following. With a slight abuse of annotation, we re-denote the optimal noise obtained by LEAD-H or LEAD-S as $\zeta _{1}^t$. Defining $\zeta _{2}^t=\frac{1}{\sigma K} \sum_{k=1}^K \delta_k^t \mathcal{F} \left( \mathcal{G}(\mathcal{Z}^t + \sigma \delta_k^t)\right)$, we further define the optimal noise as:
\begin{equation}\label{eq:lead_h_s_w}
    \delta^t_{\star} = \arg\max_{\zeta^t_{i}} \{\mathcal{F}\left(\mathcal{G}(\mathcal{Z}^t+\sigma\zeta_{i}^t)\right)\}_{i=1}^{2}.
\end{equation}
We perform LEAD guidance from the later stage of reverse diffusion process, namely, from $T_{\text{init}}$ ($\leq T$) steps. Before the guidance ($t>T_{\text{init}}$), the denoising process replicates the original one, namely
\begin{equation}
    \mathcal{A}^{t-1} \sim DDPM\left(\mathcal{D}_{1}, \mathcal{D}_{2}, \mathcal{D}_{3}, \mathcal{E}\left(\mathcal{A}^t, \mathcal{R}\right),\beta_t\right).
\end{equation}
The whole algorithm is summarized in Algorithm~\ref{Guided_DiffAb}.

\begin{table*}[ht]
\centering
\caption{\label{ddg_guided} Average AAR, RMSD, Hydro, and Pred $\Delta\Delta G$ on 19 test complexes, evaluated for each CDR. ($\Delta \Delta G$ + Hydro) refers to a two-objective optimization with equal weight. The best results w.r.t different types of guidance are highlighted in bold. ``*" denotes the difference is considered to be statistically significant with a significance level $\alpha = 0.05$.}
\renewcommand{\arraystretch}{1.2}
\setlength{\tabcolsep}{1mm}{
\resizebox{0.8\textwidth}{!}{
\begin{tabular}{lc|cccc|cccc}
\toprule[1.2pt]
Guidance  & Model    & AAR (\%) & RMSD (\AA) & Hydro& Pred $\Delta \Delta$G & AAR (\%) & RMSD (\AA) & Hydro& Pred $\Delta \Delta$G \\
&& \multicolumn{4}{c}{CDR H1}   & \multicolumn{4}{c}{CDR L1}   \\\hline
Unconditional    & DiffAb   & 65.8& 0.88  & -0.23& -0.04& 56.4& 1.22  & -0.54& -0.57    \\
\multirow{2}{*}{Pred $\Delta \Delta$G}   & GuideRaw & 65.1& 1.14  & -0.31& -0.38    & 56.5& 1.34  & -0.65& -1.28    \\
 & \cellcolor{gray!20}LEAD & \cellcolor{gray!20}38.7& \cellcolor{gray!20}0.89  & \cellcolor{gray!20}\textbf{-0.88*} & \cellcolor{gray!20}\textbf{-0.67}& \cellcolor{gray!20}31.1& \cellcolor{gray!20}1.25  & \cellcolor{gray!20}\textbf{-0.72} & \cellcolor{gray!20}\textbf{-3.11*}\\
\multirow{2}{*}{Hydro}& GuideRaw & 24.0& 0.91  & -2.93& \textbf{-0.07}& 37.4& 1.45  & -2.78& -1.14\\
& \cellcolor{gray!20}LEAD & \cellcolor{gray!20}16.1& \cellcolor{gray!20}0.92  & \cellcolor{gray!20}\textbf{-3.22*} & \cellcolor{gray!20}-0.05    & \cellcolor{gray!20}24.4& \cellcolor{gray!20}1.28  & \cellcolor{gray!20}\textbf{-3.04*} & \cellcolor{gray!20}\textbf{-1.45}\\
\multirow{2}{*}{$(\Delta \Delta$G+ Hydro)} & GuideRaw & 58.4& 1.05  & -0.78& -0.56    & 55.2& 1.24  & -0.99& -1.32    \\
& \cellcolor{gray!20}LEAD & \cellcolor{gray!20}27.0& \cellcolor{gray!20}0.86  & \cellcolor{gray!20}\textbf{-2.31*} & \cellcolor{gray!20}\textbf{-0.66}& \cellcolor{gray!20}30.7& \cellcolor{gray!20}1.26  & \cellcolor{gray!20}\textbf{-1.79*} & \cellcolor{gray!20}\textbf{-2.83*}\\
\hline
&& \multicolumn{4}{c}{CDR H2} & \multicolumn{4}{c}{CDR L2}   \\\hline
Unconditional  & DiffAb   & 49.0& 0.77  & -1.18& 0.45& 58.2& 1.06  & -0.46& -0.02    \\
\multirow{2}{*}{Pred $\Delta \Delta$G}  & GuideRaw & 47.6& 1.59  & -1.31& 0.02& 55.6& 1.76  & -0.65& -0.36    \\
& \cellcolor{gray!20}LEAD & \cellcolor{gray!20}21.5& \cellcolor{gray!20}0.70  & \cellcolor{gray!20}\textbf{-1.33} & \cellcolor{gray!20}\textbf{-1.34*}& \cellcolor{gray!20}26.1& \cellcolor{gray!20}1.34  & \cellcolor{gray!20}\textbf{-0.66} & \cellcolor{gray!20}\textbf{-1.52*}\\
\multirow{2}{*}{Hydro}    & GuideRaw & 6.10& 0.76  & -4.07& -0.24    & 22.9& 1.05  & -3.77& -0.46    \\
& \cellcolor{gray!20}LEAD & \cellcolor{gray!20}3.68& \cellcolor{gray!20}0.74  & \cellcolor{gray!20}\textbf{-4.17*} & \cellcolor{gray!20}\textbf{-0.39}& \cellcolor{gray!20}11.2& \cellcolor{gray!20}1.34  & \cellcolor{gray!20}\textbf{-4.01*} & \cellcolor{gray!20}\textbf{-0.57}\\
\multirow{2}{*}{($\Delta \Delta$G+ Hydro)}  & GuideRaw & 37.6& 1.24  & -2.17& -0.05    & 52.2& 1.30  & -1.83& -0.41    \\
& \cellcolor{gray!20}LEAD & \cellcolor{gray!20}16.4& \cellcolor{gray!20}0.73  & \cellcolor{gray!20}\textbf{-3.02*} & \cellcolor{gray!20}\textbf{-0.95*}& \cellcolor{gray!20}18.9& \cellcolor{gray!20}1.18  & \cellcolor{gray!20}\textbf{-2.68*} & \cellcolor{gray!20}\textbf{-1.45*}\\
\hline
&& \multicolumn{4}{c}{CDR H3}  & \multicolumn{4}{c}{CDR L3}  \\\hline
Unconditional    & DiffAb   & 26.3& 3.35  & -0.73& 0.14& 46.5& 1.07  & -1.31& -2.19    \\
\multirow{2}{*}{Pred $\Delta \Delta$G}& GuideRaw & 25.8& 3.41  & -1.03& -1.99    & 44.5& 1.16  & \textbf{-1.39*}& -2.94    \\
 & \cellcolor{gray!20}   LEAD & \cellcolor{gray!20}21.1& \cellcolor{gray!20}3.38  & \cellcolor{gray!20}\textbf{-1.15}& \cellcolor{gray!20}\textbf{-2.16}    & \cellcolor{gray!20}20.1& \cellcolor{gray!20}1.18  & \cellcolor{gray!20}-1.12& \cellcolor{gray!20}\textbf{-3.52*}\\
\multirow{2}{*}{Hydro}   & GuideRaw & 11.7& 3.39  & -3.87& \textbf{-1.06}& 32.2& 1.10  & -3.63& -2.04    \\
& \cellcolor{gray!20}LEAD    & \cellcolor{gray!20}8.88& \cellcolor{gray!20}3.37  & \cellcolor{gray!20}\textbf{-3.96*} & \cellcolor{gray!20}-0.94    & \cellcolor{gray!20}21.4& \cellcolor{gray!20}1.20  & \cellcolor{gray!20}\textbf{-3.87*} & \cellcolor{gray!20}\textbf{-2.22} \\
\multirow{2}{*}{($\Delta \Delta$G+ Hydro)}  & GuideRaw & 26.0& 3.63  & -1.62& -2.20    & 45.3& 1.16  & -1.93& -2.84\\
& \cellcolor{gray!20}LEAD & \cellcolor{gray!20}20.8& \cellcolor{gray!20}3.39  & \cellcolor{gray!20}\textbf{-2.26*} & \cellcolor{gray!20}\textbf{-2.35}& \cellcolor{gray!20}26.5& \cellcolor{gray!20}1.20  & \cellcolor{gray!20}\textbf{-2.28*}& \cellcolor{gray!20}\textbf{-3.25*} \\
\bottomrule[1.2pt]
\end{tabular}}}
\begin{small}
\begin{flushleft}
$^\dagger$ The results of GuideRaw are copied from the original paper, and its property guidance is incorporated from the first denoising step, i.e., $T_{\text{init}}=100$. In contrast, our LEAD adopts $T_{\text{init}}=50$, using fewer evaluation queries. 
\end{flushleft}
\end{small}\vskip-0.1in
\end{table*}

\begin{figure*}[!t]
\centering 
\centerline{\includegraphics[width=\textwidth]{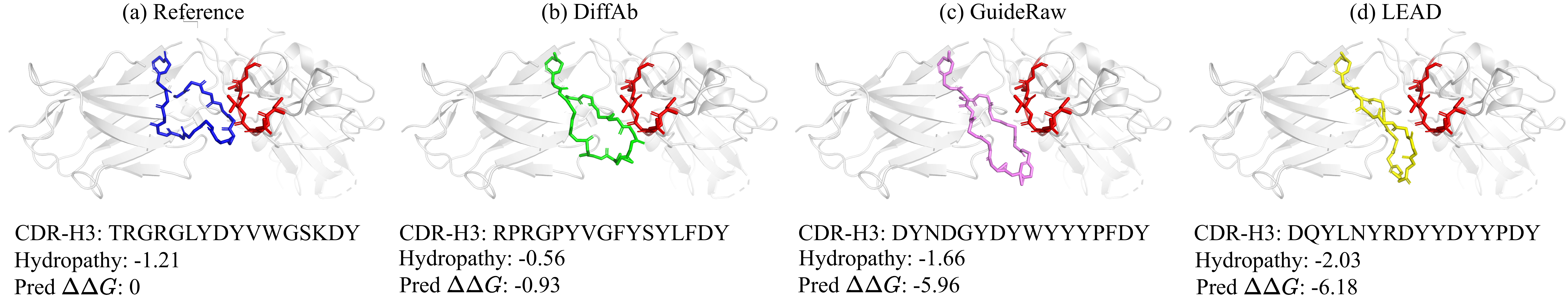}}
\vskip-.01in\caption{\label{fig:vis} Visualization of sequence-structure designs generated by various methods for optimizing Pred $\Delta\Delta G$. The presented test complex is 7chf\_A\_B\_R. The antigen epitope is displayed in red color. The other colored one is the designed CDR-H3.} \vskip-.05in
\end{figure*}

\section{Experiment}
We follow \cite{Amelia2024guiding} to consider two developability properties, i.e., hydropathy score and folding energy for antibody design.
\begin{itemize}[left=0pt, labelsep=1em]
\item Hydropathy score (abbr. Hydro). It averages the hydropathy values over the generated CDR sequences and serves as a proxy for solubility and aggregation~\cite{kyte1982simple}. Negative scores indicate hydrophilicity while positive scores indicate hydrophobicity, and small score is preferred.   
\item Folding energy difference (abbr. $\Delta\Delta G$). It denotes the structural stability of generated antibodies relative to the reference. Following~\cite{Amelia2024guiding}, we predict $\Delta\Delta G$ (abbr. Pred $\Delta\Delta G$) using an existing predictor released by~\cite{shan2022deep}, which takes as input the amino acid sequence and $\mathrm{C}_\alpha$ atom coordinates. A lower value suggests improved structural stability.  
\end{itemize}

\paragraph{Dataset} 
We utilize the test set described in~\cite{luo2022antigen}, which consists of 19 antibody-antigen complexes derived from the SabDab database~\cite{dunbar2014sabdab}. The test set includes protein antigens from various pathogens, such as influenza and SARS-CoV-2.

\begin{figure*}[!t]
\centering 
\centerline{\includegraphics[width=.9\textwidth]{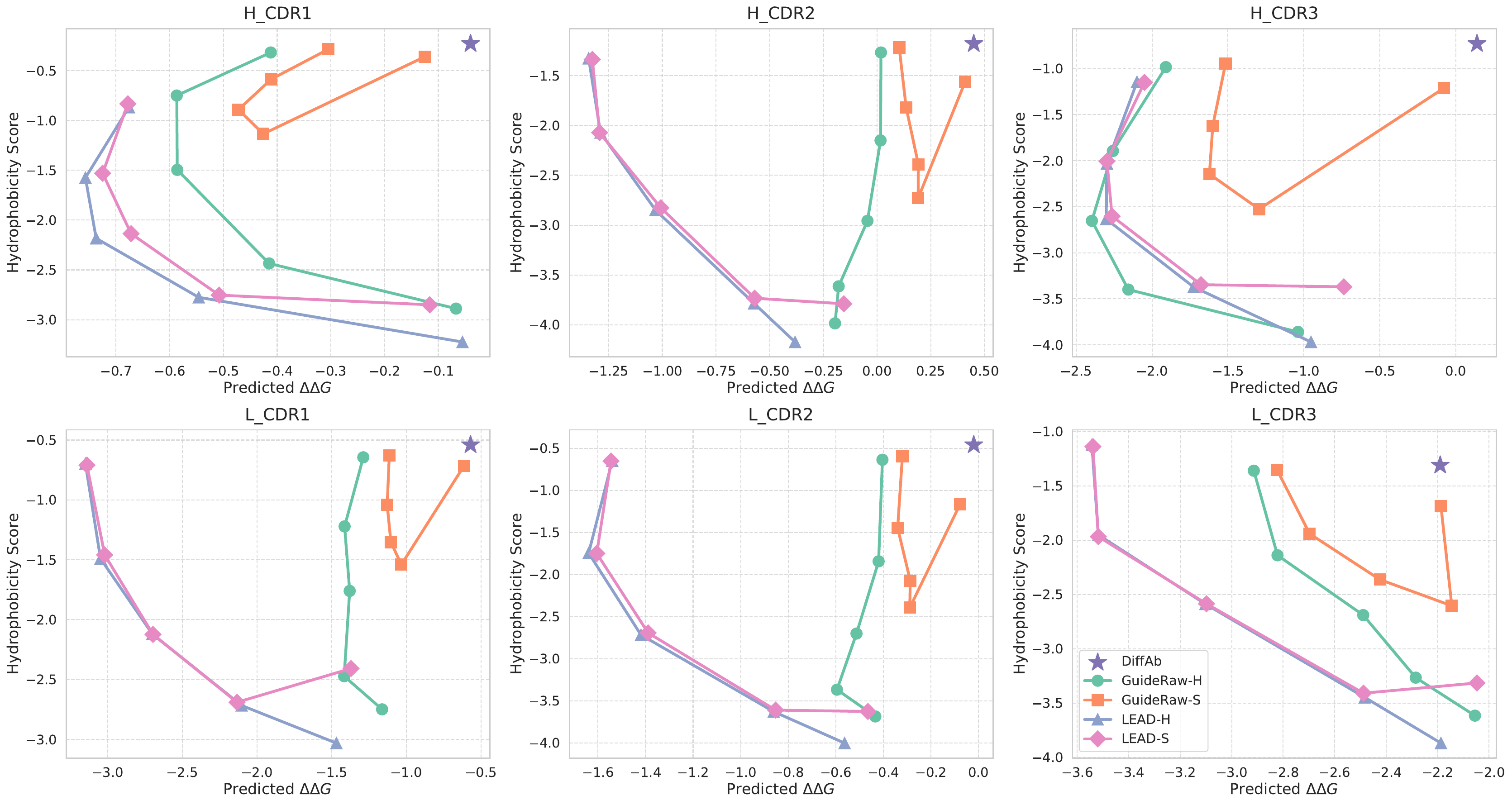}}\vskip-0.1in 
\caption{\label{fig:moo} Trade-off in multi-property optimization. Each point represents the average Hydro and Pred $\Delta\Delta G$ on 19 test complexes, evaluated for each CDR. Different weights are used to combine the two objectives, i.e., $w*\Delta\Delta G + (1-w)*\text{Hydro}$, with $w = 0, 0.25, 0.5, 0.75, 1$.} 
\vskip-0.1in
\end{figure*}

\paragraph{Hyperparameters}
Unless otherwise specified, the latent noise scale~$\sigma$ is adaptively set to $\beta_t$ (as in Eq.~\eqref{denoising_independent}) and the number of evaluation per time step~$K$ is set to 20, the initial guidance step~$T_{\text{init}}$ is set to 50. \revision{It is more effective to add guidance at a later stage of the denoising process as samples from the earlier steps can be too noisy to evaluate accurately.}

\paragraph{Baselines}
We employ the widely adopted antibody sequence-structure co-design diffusion model framework, DiffAb~\cite{luo2022antigen} as the pretrained model. We consider the relevant work, property optimization under the framework of sequence-structure co-design as baselines\footnote{Other baselines DIFFFORCE~\cite{kulyte2024improving} and ABDPO~\cite{zhou2024antigen} do not release their codes. Nevertheless, DIFFFORCE requires a differential force field model, which is inapplicable to the black-box setting of this work. ABDPO needs to fine-tunes DIFFAB, requiring a large amount of property evaluations.}. \cite{Amelia2024guiding} performs guidance similar to our hard/soft selection strategy, but on the raw data space. We name it as GuideRaw-H and GuideRaw-S for hard and soft selection version, respectively. Except where mentioned, the experimental setup follows the setting of the original work. That is, the initial guidance step~$T_{\text{init}}$ is set to 100 and the number of evaluation per time step~$K$ is set to~20.

\paragraph{Evaluation}
In addition to the properties mentioned above, we evaluate these designs using the following metrics: (i) \textbf{Amino Acid Recovery (AAR)}, which measures the sequence identity between the reference and generated CDR sequences; (ii) \textbf{Root Mean Square Deviation (RMSD)}, which computes the $C_{\alpha}$ atom distance between the reference and generated CDR structures. \revision{AAR and RMSD are reported for reference purposes and are not bolded in Table results. For a detailed explanation, please refer to point (4) in Section~\ref{sect:4.1}.}

\subsection{Single Property Optimization}\label{sect:4.1}
For all methods, we generate 100 designs per CDR (six CDRs in total) for each of the 19 test antigen-antibody complexes. The guidance method is to optimize the hydropathy score and the predicted $\Delta\Delta G$, respectively. Here, both GuideRaw and our LEAD adopt the hard selection version for a fair comparison. Other guidance strategies are studied in Section~\ref{guidance_strategy}.

We report the evaluation results in Table~\ref{ddg_guided}.
(1) The results show that our LEAD outperforms GuideRaw in terms of both target properties, which is attributed to our optimization over the shared latent code ensuring the synchronization of different modalities.
(2) LEAD conducts guidance in only 50 steps, compared to the 100 steps required by GuideRaw. This demonstrates that our method achieves better performance while requiring fewer black-box function evaluations.
(3) DiffAb performs unconditional generation and exhibits strong AAR and RMSD scores. This indicates that the antibody generated by this sequence-structure co-design model can recover CDR structures similar to the given reference. However, the target properties of the antibodies produced by DiffAb fall below expectations.
(4) Property optimization often leads to a degradation of AAR and RMSD, as it requires adjusting the generated CDRs to meet property requirements. Specifically, the hydropathy score is determined by the sequences only. As a result, we observe that antibodies optimized by all property guidance methods exhibit significant AAR degradation but improved hydropathy scores. Notably, the RMSD scores remain relatively low across all generations, indicating that the generated CDR structures are still similar to the references.

Fig.~\ref{fig:vis} visualizes the generated antibodies. It shows that all methods can generate valid sequences and structures. GuideRaw and our LEAD obtain better hydropathy score and predicted $\Delta\Delta G$ than natural antibodies.

\begin{figure*}[!t]
\centering
\begin{minipage}{.33\linewidth}
\centerline{\includegraphics[width=1\textwidth]{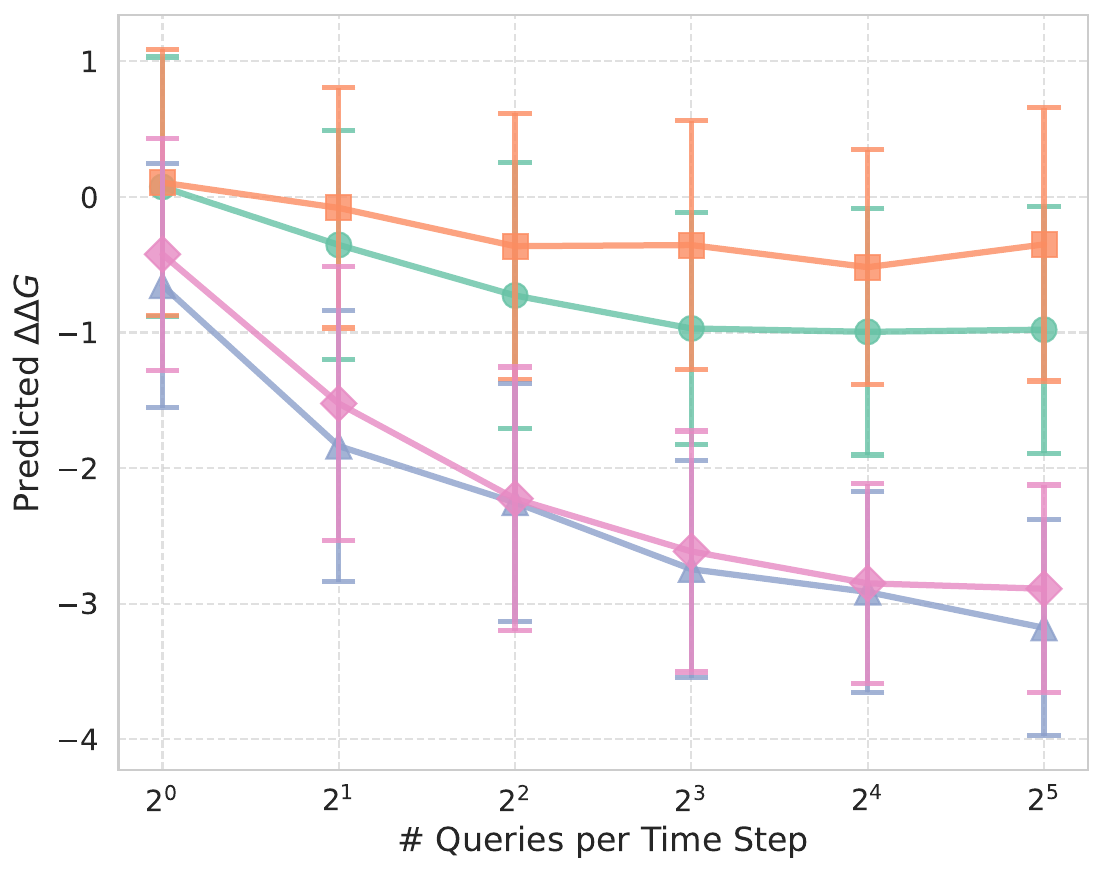}}
\vskip-0.05in \centerline{\footnotesize (a)  Predicted $\Delta\Delta$G guided antibody design}
\end{minipage} 
\hspace{0.3in}
\begin{minipage}{.33\linewidth}
\centerline{\includegraphics[width=1\textwidth]{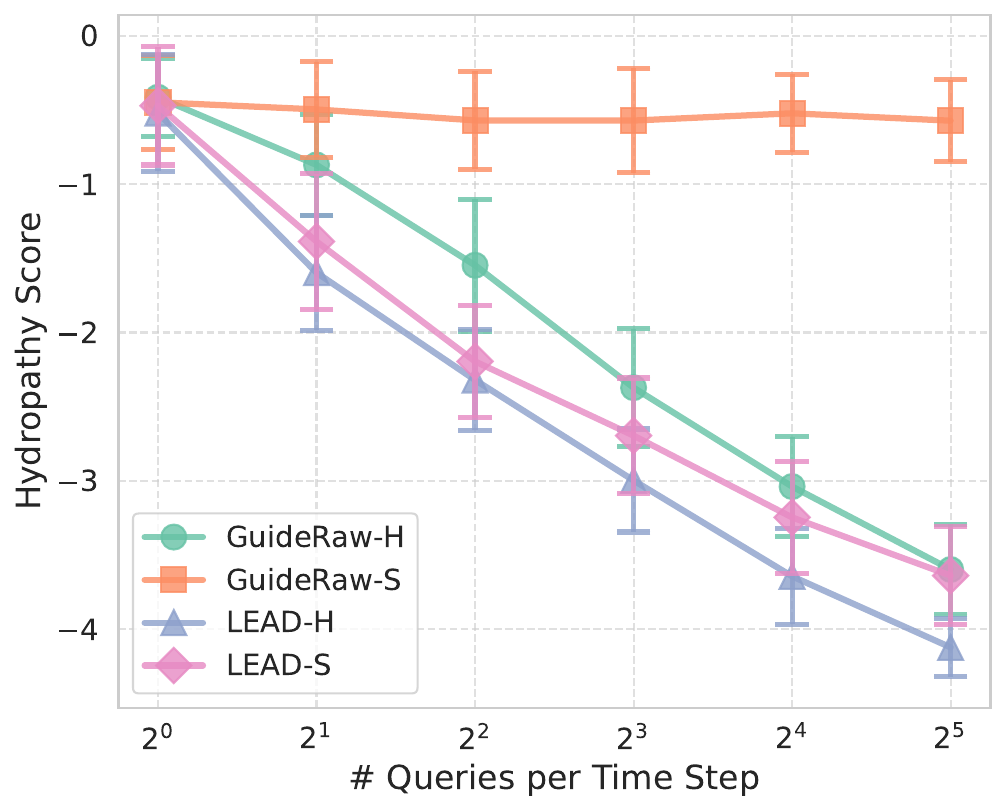}}
\vskip-0.05in \centerline{\footnotesize (b)  Hydropathy score guided antibody design}
\end{minipage} 
\vskip-0.05in
\caption{\label{fig:query_efficient} Query-efficient comparison on the test complex 5xku\_C\_B\_A (design H-CDR1) between GuideRaw and our LEAD w.r.t. two selection strategies. For a fair comparison, property guidance is incorporated for all methods after 50 time steps respectively.} \vskip-0.05in 
\end{figure*}

\begin{table*}[ht]
\centering
\caption{\label{combined} Average AAR, RMSD, Hydro, and Pred $\Delta\Delta G$ on the test complex 5xku\_C\_B\_A (design H-CDR1). For a fair comparison, property guidance is incorporated for all methods after 50 time steps respectively.}\vskip-.05in
\renewcommand{\arraystretch}{1.2}
\setlength{\tabcolsep}{1mm}{
\resizebox{0.8\textwidth}{!}{
\begin{tabular}{c|cccc|cccc}
\toprule[1.2pt]
Guidance & \multicolumn{4}{c|}{Pred $\Delta\Delta$G} & \multicolumn{4}{c}{Hydro} \\
Metric & AAR (\%) & RMSD (\AA) & Hydro & Pred $\Delta\Delta$G & AAR (\%) & RMSD (\AA) & Hydro &  Pred $\Delta\Delta$G \\
\midrule[1pt]
DiffAb & 58.4 $\pm$   13.9 & 1.28 $\pm$ 0.20 & -0.43 $\pm$ 0.39 & -0.75 $\pm$ 0.87 & 58.4 $\pm$   13.9 & 1.28 $\pm$ 0.20 & -0.43 $\pm$ 0.39 & -0.75 $\pm$ 0.87 \\\hline
GuideRaw-H & 66.1 $\pm$ 11.2 & 1.62 $\pm$ 0.40 & -0.57 $\pm$ 0.34 & -1.01 $\pm$ 0.93 & 15.2 $\pm$ 12.3 & 1.37 $\pm$ 0.29 & -3.25 $\pm$ 0.28 & -0.99 $\pm$ 0.79 \\
GuideRaw-S & 66.8 $\pm$ 12.2 & 1.53 $\pm$ 0.33 & -0.50 $\pm$ 0.35 & -0.81 $\pm$ 0.97 & 68.4 $\pm$ 10.8 & 1.39 $\pm$ 0.29 & -0.59 $\pm$ 0.31 & -0.11 $\pm$ 0.91 \\\hline
LEAD-H & 13.3 $\pm$ 5.12 & 1.35 $\pm$ 0.24 & \textbf{-1.10 $\pm$ 0.56} & -2.66 $\pm$ 0.69 & 4.78 $\pm$ 5.84 & 1.33 $\pm$ 0.24 & \underline{-3.75 $\pm$ 0.25} & -1.73 $\pm$ 0.76 \\
LEAD-S & 12.2 $\pm$ 2.29 & 1.35 $\pm$ 0.23 & -0.83 $\pm$ 0.56 & -2.65 $\pm$ 0.71 & 10.4 $\pm$ 9.05 & 1.33 $\pm$ 0.27 & -3.22 $\pm$ 0.33 & -1.65 $\pm$ 0.84 \\
LEAD-W & 37.0 $\pm$ 17.4 & 1.67 $\pm$ 0.21 & -0.28 $\pm$ 0.62 & -2.64 $\pm$ 0.93 & 35.1 $\pm$ 17.2 & 1.47 $\pm$ 0.26 & -0.77 $\pm$ 0.44 & -1.21 $\pm$ 0.93\\
 \rowcolor{gray!20}LEAD-(W+H) &14.1 $\pm$ 7.70&1.43 $\pm$ 0.23&\underline{-0.97 $\pm$ 0.55}&\underline{-2.89 $\pm$ 0.81}&3.80 $\pm$ 3.72&1.30 $\pm$ 0.26&\textbf{-3.78 $\pm$ 0.23}& \underline{-1.78 $\pm$ 0.70}\\
\rowcolor{gray!20}LEAD-(W+S)&14.4 $\pm$ 8.48&1.45 $\pm$ 0.24&-0.83 $\pm$ 0.65&\textbf{-2.90 $\pm$ 0.80}&7.34 $\pm$ 6.48&1.29 $\pm$ 0.21&-3.38 $\pm$ 0.34&\textbf{-1.88 $\pm$ 0.68}\\
\bottomrule[1.2pt]
\end{tabular}}}\vskip-.05in
\end{table*}

\subsection{Multi-Property Optimization}
\revision{Optimizing multiple properties presents challenges because each property can have different value scales and units. To address this, we normalize property values to ensure they contribute comparably during optimization. We adopt a weighted sum with adjustable weights to balance trade-offs according to design priorities.}
Here, we optimize both the hydropathy score and $\Delta\Delta G$ with $w*\Delta\Delta G + (1-w)*\text{Hydro}$. The weight $w$ is set as $0, 0.25, 0.5, 0.75, 1$, respectively. For a fair comparison, we set up the same number of function queries for our LEAD and GuideRaw in this setting. That is, both methods set $T_{\text{init}}=50$ and $K=20$. Still, we generate 100 generated designs per CDR for each of the 19 test antigen-antibody complexes. 

The results are summarized in Fig.~\ref{fig:moo}. The hydropathy and energy properties exhibit a trade-off, where improving one often comes at the expense of the other. Our LEAD effectively navigates this balance, achieving superior solutions compared to the baseline approaches.

\subsection{Query Efficiency}
By guiding optimization in the shared latent space, our LEAD captures property semantics effectively and preserves the correspondence between sequence and structure, leading to more efficient guidance. We conduct experiments with different numbers of queries, $2^0, 2^1, 2^2, 2^3, 2^4, 2^5$ per guided denoising step. In this experiment, all methods generate 100 designs for H-CDR1 from one of the test complexes and perform property-guided denoising after 50 denoising steps. 

The results in Fig.~\ref{fig:query_efficient} demonstrate the efficiency of our method. To be specific, with an increased number of black-box evaluations, the optimization process explores a wider range of possibilities, leading to a more thorough refinement of the target property.
It is interesting to see that our LEAD achieves a much larger performance gain in optimizing the predicted $\Delta\Delta G$ than in optimizing the hydropathy score. The predicted $\Delta\Delta G$ is relevant to both sequences and structures, while the hydropathy score is only determined by sequences. This particularly demonstrates our method's superiority on those sequence-structure derived properties.

\subsection{Guidance Strategies}
\label{guidance_strategy}
Here, we explore the efficacy of different guidance strategies demonstrated in Section~\ref{reward_guided}, namely, LEAD-H, LEAD-S, LEAD-W, LEAD-(W+H), LEAD-(W+S). 

(1) Our LEAD with different guidance strategies achieves better results in terms of optimizing predicted~$\Delta\Delta G$. This property is relevant to both antibody sequence and structure. The guidance over the shared latent space indeed enhances its optimization. (2) In terms of the hydropathy score, LEAD-W does not achieve efficient optimization, resulting in a score similar to that of unconditional generation. This is because the hydropathy score is a sequence derived property, and its gradient space is not continuous. As a result, the weighted strategy fails to perform an effective search. Instead, the discrete search strategies, LEAD-H and LEAD-S, are effective in this case. (3) The weighted strategy (Eq.~\eqref{eq:lead_h_s_w}) can slightly boost the discrete search strategies, as demonstrated by the results of LEAD-(W+H) and LEAD-(W+S).

\section{Conclusion}
\revision{In this work, we introduce LEAD, a black-box guidance framework that operates in a shared latent space of sequence and structure, enabling efficient property optimization for antibody CDRs without model retraining. LEAD is generally effective at maintaining sequence-structure synchronization; however, its performance is limited by the capacity of DiffAb, which may fail to preserve meaningful alignment between sequence and structure. More advanced co-design frameworks~\cite{yang2025repurposing,uccar2024benchmarking} may offer improved capabilities in this regard. Future work will also incorporate wet-lab experiments to validate computational designs and provide feedback to further guide optimization. Beyond antibody design, LEAD is broadly applicable to other AI-driven co-design tasks involving coupled sequence-structure modalities, e.g., protein, RNA, and small-molecule design.}


\section*{Acknowledgments} 
This paper was supported by the Manufacturing Trade and Connectivity (MTC) Individual Research Grant (IRG) (Grant No. M24N7c0091). It is also supported by the National Research Foundation, Singapore under its National Large Language Models Funding Initiative (AISG Award No: AISG-NMLP-2024-004). Any opinions, findings and conclusions or recommendations expressed in this material are those of the author(s) and do not reflect the views of National Research Foundation, Singapore.

\bibliographystyle{named}
\bibliography{ijcai25}

\clearpage

\appendix

\section{Additional Experiments}
We additionally consider the likelihood as the guidance, which can enhance data quality for antibody design~\cite{gruver2024protein}.
\begin{itemize}
    \item ProtGPT negative log-likelihood (abbr. NLL). This score is assigned by ProtGPT~\cite{ferruz2022protgpt2}, a language model trained on Uniref50~\cite{suzek2007uniref}, denoting the naturalness of generated antibody sequences, the lower the better.
\end{itemize}

\begin{figure}[!ht]
\centering
\begin{minipage}{.97\linewidth}
\centerline{\includegraphics[width=1\textwidth]{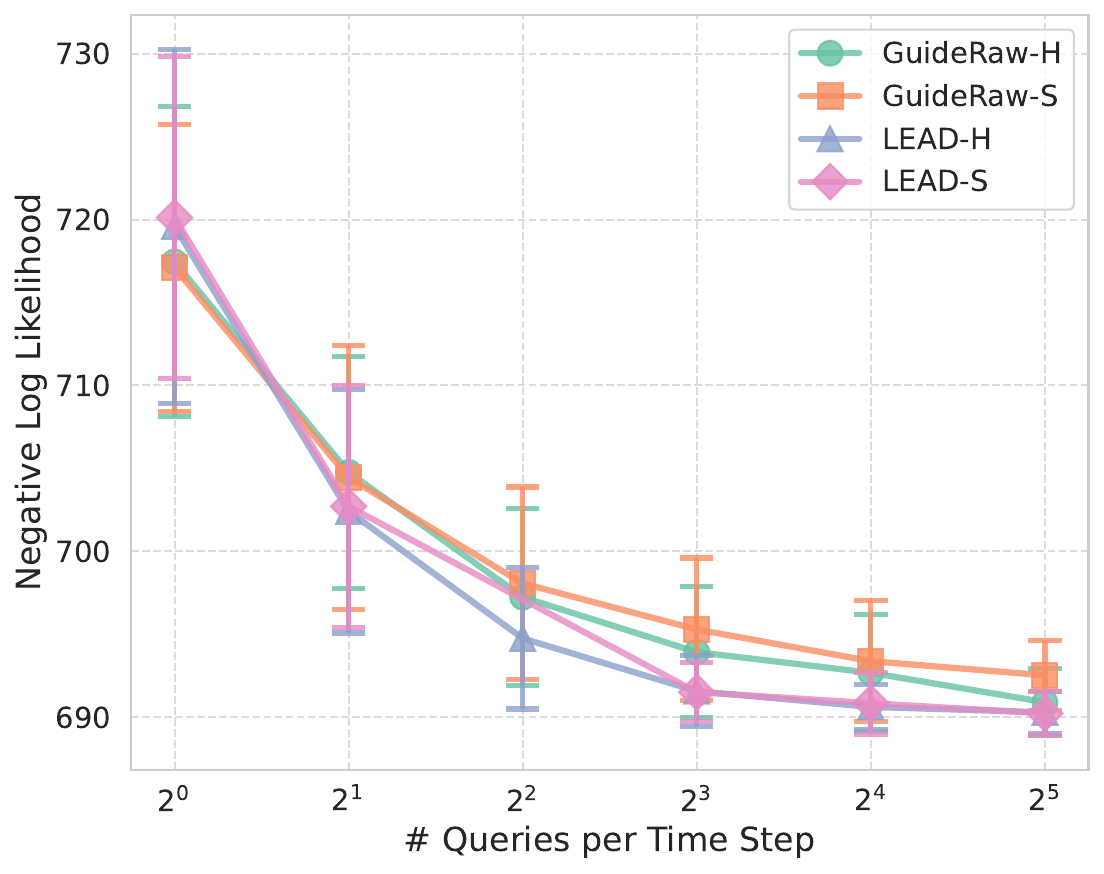}}
\end{minipage} \vskip-0.05in
\caption{\label{fig:query_efficient_nll}  Query-efficient comparison on the test complex 5xku\_C\_B\_A (design H-CDR1) between GuideRaw and our LEAD w.r.t. two selection strategies. For a fair comparison, property guidance of NLL is incorporated for all methods after 50 time steps respectively.} 
\end{figure}

Fig.~\ref{fig:query_efficient_nll} presents the experiments with varying numbers of queries. Similar to the results in Fig.~\ref{fig:query_efficient} of the main text, our LEAD method achieves better performance with fewer queries (half or even a quarter) in the NLL-guided optimization.

As shown in Table~\ref{combined_nll}, with sufficient function queries ($K=20$), LEAD and GuideRaw achieve similar performance in the optimization of NLL. Moreover, the weighted strategy of LEAD fails in this case, similar to the results observed for the hydropathy score in the right panel of Table~\ref{combined} (in the main text). This occurs because the property is derived exclusively from the sequence, resulting in a discrete gradient space.

\begin{table}[!ht]
\centering
\caption{\label{combined_nll} Average AAR, RMSD, Hydro, and Pred $\Delta\Delta G$ on the test complex 5xku\_C\_B\_A (design H-CDR1). The best results are highlighted in bold while the second best results are marked with an underline. For a fair comparison, property guidance  of NLL is incorporated for all methods after 50 time steps respectively.}
\renewcommand{\arraystretch}{1.2}
\setlength{\tabcolsep}{1mm}{
\resizebox{0.5\textwidth}{!}{
\begin{tabular}{c|ccccc}
\toprule[1.2pt]
Metric & AAR (\%) & RMSD (\AA) & Hydro & Pred $\Delta\Delta$G & NLL ($\times 10^2$) \\\midrule[1pt]
DiffAb & 62.2 $\pm$   12.1 & 1.24 $\pm$ 0.19 & \underline{-0.48 $\pm$ 0.32} & -0.65 $\pm$ 0.91 & -7.20 $\pm$ 0.11 \\\hline
GuideRaw-H & 58.4 $\pm$ 7.836 & 1.34 $\pm$ 0.28 & -0.39 $\pm$ 0.24 & -0.94 $\pm$ 0.86 & -6.92 $\pm$ 0.03 \\
GuideRaw-S & 62.5 $\pm$ 10.1 & 1.33 $\pm$ 0.23 & \textbf{-0.50 $\pm$ 0.28} & -0.87 $\pm$ 0.96 & -6.93 $\pm$ 0.03 \\\hline
LEAD-H & 54.9 $\pm$ 10.1 & 1.28 $\pm$ 0.23 & -0.33 $\pm$ 0.25 & -1.25 $\pm$ 0.81 & -6.91 $\pm$ 0.02 \\
LEAD-S & 56.4 $\pm$ 10.6 & 1.33 $\pm$ 0.22 & -0.36 $\pm$ 0.24 & \textbf{-1.38 $\pm$ 0.8} & -6.91 $\pm$ 0.01 \\
LEAD-W & 41.3 $\pm$ 17.7 & 1.42 $\pm$ 0.22 & -0.26 $\pm$ 0.49 & -1.33 $\pm$ 0.94 & -7.20 $\pm$ 0.11 \\
\rowcolor{gray!20}LEAD-(W+H) & 56.8 $\pm$ 11.5 & 1.28 $\pm$ 0.23 & -0.36 $\pm$ 0.17 & -1.28 $\pm$ 0.79 & \textbf{-6.90 $\pm$ 0.01} \\
\rowcolor{gray!20}LEAD-(W+S) & 55.17 $\pm$ 12.0 & 1.33 $\pm$ 0.21 & -0.33 $\pm$ 0.2 & \underline{-1.36 $\pm$ 0.85} & \textbf{-6.90 $\pm$ 0.01} \\
\bottomrule[1.2pt]
\end{tabular}}}
\end{table}

\section{More Details}
The amino acid of the antigen epitope for ``7chf\_A\_B\_R" is ``CNGVEG"~\cite{sars2020}, which is colored in red in Fig.~2 of the main text.

The algorithms of DiffAb and GuideRaw can be seen in Algorithm~\ref{DiffAb} and Algorithm~\ref{Best_N_DiffAb}.

\begin{algorithm}[!hb]
        \caption{Unconditional antibody generation}\label{DiffAb}
        \begin{algorithmic}[1]
          \STATE {\bfseries Input:} a pre-trained antibody diffusion model featuring three-output denosing network with a shared encoder~($\mathcal{E}, \mathcal{D}_{1}, \mathcal{D}_{2}, \mathcal{D}_{3}$),  the antibody-antigen backbone $\mathcal{R}$, the dynamic step size~$\{\beta_t\}_{t=1}^T$.
          \STATE {\bfseries Initialize:} $\mathcal{A}^T= \big(\mathbf{s}^T\sim \mathcal{C}\left(\mathbb{I}(\mathbf{s}^T) |\mathbf{1}/20 \right), \mathbf{x}^T\sim \mathcal{N}(\mathbf{x}^T|\mathbf{0}, \mathbf{I}), \mathbf{O}^T_i \sim \mathcal{I} \mathcal{G}_{\mathrm{SO}(3)}\left(\mathbf{I}, 1\right)\big)$.
          \FOR {$t = T,T-1,\ldots, 1$}
          \STATE $\mathcal{A}^{t-1} \sim DDPM\left(\mathcal{D}_{1}, \mathcal{D}_{2}, \mathcal{D}_{3}, \mathcal{E}\left(\mathcal{A}^t, \mathcal{R}\right),\beta_t\right)$;
          \ENDFOR
          \STATE {\bfseries Output:} generated CDR loop $\mathcal{A}^0$.
        \end{algorithmic}
\end{algorithm}

\begin{algorithm}[!hb]
        \caption{Guided antibody generation in raw data space}\label{Best_N_DiffAb}
        \begin{algorithmic}[1]
        \STATE {\bfseries Input:} a pre-trained antibody diffusion model featuring three-output denosing network with a shared encoder~($\mathcal{E}, \mathcal{D}_{1}, \mathcal{D}_{2}, \mathcal{D}_{3}$), the antibody-antigen backbone $\mathcal{R}$, the dynamic step size~$\{\beta_t\}_{t=1}^T$, the property evaluator~$\mathcal{F} (\uparrow)$, number of evaluations per time step $K$.
          \STATE {\bfseries Initialize:} $\mathcal{A}^T= \big(\mathbf{s}^T\sim \mathcal{C}\left(\mathbb{I}(\mathbf{s}^T) |\mathbf{1}/20 \right), \mathbf{x}^T\sim \mathcal{N}(\mathbf{x}^T|\mathbf{0}, \mathbf{I}), \mathbf{O}^T_i \sim \mathcal{I} \mathcal{G}_{\mathrm{SO}(3)}\left(\mathbf{I}, 1\right)\big)$.
          \FOR {$t = T,T-1,\ldots, 1$}
          \FOR {$k = 1,2,\ldots, K$}
          \STATE $\mathcal{A}^{t-1}_k \sim DDPM\left(\mathcal{D}_{1}, \mathcal{D}_{2}, \mathcal{D}_{3}, \mathcal{E}\left(\mathcal{A}^t, \mathcal{R}\right),\beta_t\right)$;
         \ENDFOR
         \STATE $\mathcal{A}^{t-1}_{\star} = \underset{\mathcal{A}^{t-1}_k}{\arg\max}\{\mathcal{F}(\mathcal{A}^{t-1}_k)\}_{k=1}^K$;
          \ENDFOR
          \STATE {\bfseries Output:} generated CDR loop $\mathcal{A}^0_{\star}$.
        \end{algorithmic}
\end{algorithm}

\end{document}